\documentclass{article}

\usepackage{microtype}
\usepackage{graphicx}
\usepackage{subfigure}
\usepackage{booktabs} % for professional tables

\usepackage{hyperref}
\usepackage{multirow}

\usepackage[accepted]{icml2024}

\usepackage{amsmath}
\usepackage{amssymb}
\usepackage{mathtools}
\usepackage{amsthm}

\usepackage[capitalize,noabbrev]{cleveref}

\theoremstyle{plain}

\theoremstyle{definition}

\theoremstyle{remark}

\usepackage[textsize=tiny]{todonotes}

\newcommand{\framework}{\textsc{Spunge}}

\newcommand{\attr}{\mathtt{attr}}

\newcommand{\RMU}{\texttt{RMU}}
\newcommand{\TVN}{\texttt{TVN}}
\newcommand{\TIES}{\texttt{TIES}}
\newcommand{\proc}{\mathtt{proc}}
\newcommand{\zephyr}{\textsc{Zephyr-7b-beta}}
\newcommand{\llama}{\textsc{Llama2-7b}}
\icmltitlerunning{Split, Unlearn, Merge}

\begin{document}

\twocolumn[
\icmltitle{Split, Unlearn, Merge: \\ Leveraging Data Attributes for More Effective Unlearning in LLMs}

% SPUNGE:

%##########
% SPUR:

% Split-then-Merge: A  Framework for Empowering Unlearning in LLMs
% Split-then-Merge: A Framework for Fortifying Unlearning in LLMs

% Split-then-Merge: Unlearning Toxicity and Hate Speech in LLMs
% Split, Unlearn, then Merge: Unlearning Toxicity and Hate Speech in LLMs
% SPUR: Leveraging Data Attributes for Unlearning Toxicity in LLMs
% Split, Unlearn, then Merge: Unlearning Toxicity and Hate Speech in LLMs



\icmlsetsymbol{equal}{*}

\begin{icmlauthorlist}
\icmlauthor{Swanand Ravindra Kadhe}{yyy}
\icmlauthor{Farhan Ahmed}{yyy}
\icmlauthor{Dennis Wei}{yyy}
\icmlauthor{Nathalie Baracaldo}{yyy}
\icmlauthor{Inkit Padhi}{yyy}
\end{icmlauthorlist}

\icmlaffiliation{yyy}{IBM Research}

\icmlcorrespondingauthor{Swanand Ravindra Kadhe}{Swanand.Kadhe@ibm.com}

% \icmlkeywords{Unlearning, Safety, LLM}

\vskip 0.3in
]



\printAffiliationsAndNotice{}  % leave blank if no need to mention equal contribution

\begin{abstract}
Large language models (LLMs) have shown to pose social and ethical risks such as generating toxic language or facilitating malicious use of hazardous knowledge. 
Machine unlearning is a promising approach to improve LLM safety by directly removing harmful behaviors and knowledge. 
% However, prior unlearning techniques do not take into account attributes of the data used for unlearning harmful behaviors. 
In this paper, we propose ``\textbf{SP}lit, \textbf{UN}learn, Mer\textbf{GE}'' (\framework), a framework that can be used with any unlearning method to amplify its effectiveness. 
\framework{} leverages data attributes during unlearning by splitting unlearning data into subsets based on 
% judiciously selected attributes, 
specific attribute values, 
unlearning each subset separately, and merging the unlearned models. 
We empirically demonstrate that \framework{} significantly improves the performance of two recent unlearning methods on state-of-the-art LLMs while maintaining their general capabilities on standard academic benchmarks. 

% Several methods have recently been proposed for efficiently unlearning harmful behaviors or knowledge. In this paper, we demonstrate that it is possible on We propose a , and improve LLM safety. is gaining increasing attention for directly improving the safety of LLMs.  is gaining atten  ) and contain hazardous contain hazardous knowledge which canThis document provides a basic paper template and submission guidelines.
% Abstracts must be a single paragraph, ideally between 4--6 sentences long.
% Gross violations will trigger corrections at the camera-ready phase.
% \warning{this paper contains content generated by language models that may be offensive or upsetting to some readers.}
\end{abstract}

\section{Introduction}
\label{intro}
% \TODO{general sentence about harms of LLMs, to make this less abrupt.}
The rapid improvement and increasing adoption of large language models (LLMs) has been accompanied by their downsides, notably their potential harmful behaviors  \cite{weidinger2022taxonomy}. %and enable malicious actors.
LLMs are known to generate harmful content such as toxic, offensive, or hateful language \cite{sheng2019woman,gehman2020realtoxicityprompts}. LLMs also contain hazardous knowledge of sensitive topics such as biosecurity and cybersecurity, which can be (mis)used to empower malicious actors \cite{sandbrink2023artificial,fang2024llm}. 
A widely adopted way to safeguard against harmful or objectionable responses is to \textit{align} LLMs via fine-tuning %to ensure that the models refuse to answer provocative queries 
\cite{ouyang2022training,bai2022constitutional,korbak2023pretraining,glaese2022improving}.
However, current approaches such as reinforcement learning with human feedback (RLHF) are computationally expensive and have shown to be vulnerable to adversarial or \textit{jailbreak} attacks where adversarial prompts break through alignment and re-invoke harmful responses \cite{wei2023jailbroken, zou2023universal, carlini2023are}. Even subsequent benign fine-tuning can degrade alignment \cite{qi2024finetuning}.

% More challenging is implicit toxicity \cite{hartvigsen2022toxigen}. Recent LLMs such as Llama-2 \cite{touvron2023llama} and Orca \cite{mukherjee2023orca} have started to measure implicit toxicity on Toxigen.

% 

In parallel, machine \emph{unlearning} has emerged as a promising paradigm for more targeted and efficient sociotechnical harm reduction. It has been shown that unlearning can reduce toxicity and other harmful responses \cite{ilharco2023editing,zhang2023composing,yao2024large} and erase hazardous scientific knowledge \cite{li2024wmdp}. Unlearning can be considered a complementary safety tool to alignment techniques and can be used before or after alignment \cite{liu2024rethinking}.
Prior work on unlearning in LLMs has focused on developing efficient unlearning methods, 
% without paying much attention to how the unlearning data is treated 
without taking into account characteristics of unlearning data
\cite{xu2023survey,liu2024rethinking} (see Appendix~\ref{sec:related-work}).

\begin{figure}[tb]
% \vskip 0.2in
\begin{center}
\centerline{\includegraphics[width=\columnwidth]{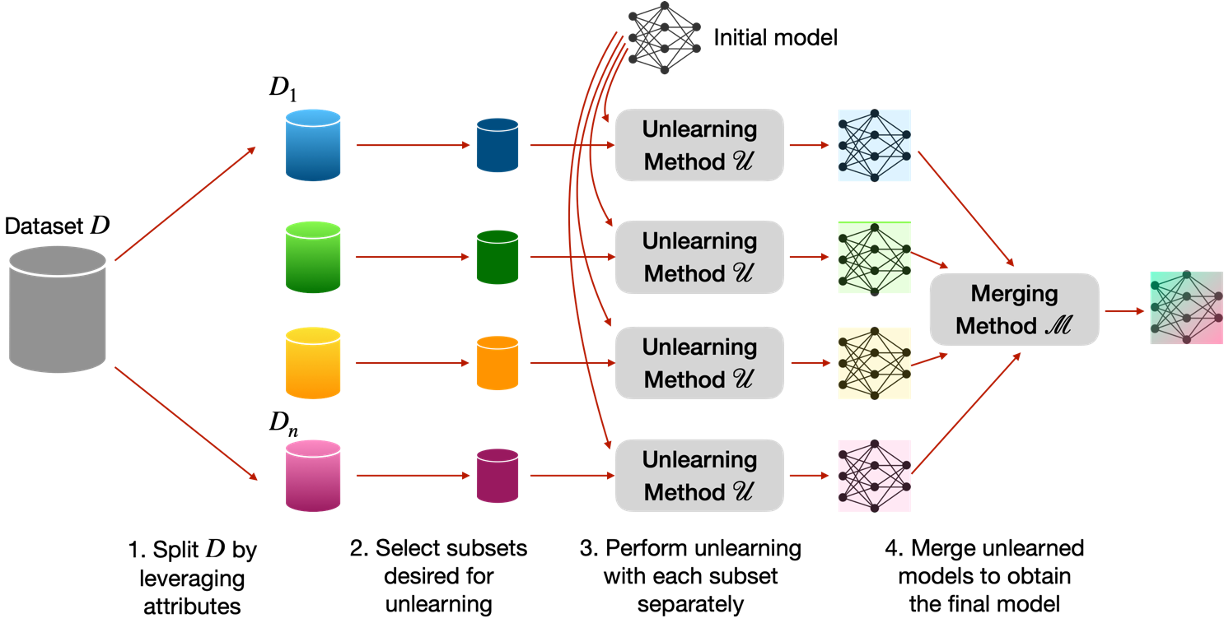}}
\caption{An Overview of the SPlit, UNlearn, then merGE (\framework) Framework. \framework{} splits the unlearning dataset into subsets based on selected attribute values, unlearns each subset separately, and then merges the unlearned models.}
\label{fig:framework}
\end{center}
\vskip -0.3in
\end{figure}

In this work, we demonstrate that leveraging \emph{attributes} in the unlearning data can significantly improve the effectiveness of unlearning. %In order to leverage data attributes in a systematic and principled manner, 
We propose a simple yet effective framework, \framework: ``\textbf{SP}lit, \textbf{UN}learn, then mer\textbf{GE}'' which operates in three steps (see Figure~\ref{fig:framework}): (i) the unlearning data is split into subsets based on the values of a selected attribute; (ii) each subset is separately used to unlearn a subtype of the undesired behavior, resulting in multiple unlearned LLMs; (iii) the unlearned LLMs are \textit{merged} to obtain the final unlearned LLM. \framework\ can be used with any unlearning method to potentially improve its effectiveness without impacting the LLM's general performance for other tasks.

% \framework\ draws its inspiration from the emerging area of \textit{model merging} in multi-task learning \cite{ilharco2023editing,matena2022merging,yadav2023ties,yu2024language}. \framework{} differs in splitting what is usually considered a single unlearning task into multiple sub-tasks, with each sub-task focused on a specific attribute. For instance, for unlearning toxicity, a sub-task can be unlearning toxicity against women. 
%% For unlearning hazardous knowledge, a subs-task can be unlearning cybersecurity  hazardous knowledge.  
% \framework{} then merges the unlearned models to combine their specialties. 
% \textcolor{blue}{I don't think the reader gets much information about this, and it does make it seem less novel. What about moving it to a later stage?}
% \TODO{sub-task as unlearning specific attributes such as demographics} \TODO{DW: don't think we need more than a sentence or two on the multi-task inspiration}

\textbf{Our Contributions:} 
%\TODO{in progress} %We outline our contributions in the following. 

\begin{itemize}
    \item We propose the \framework{} framework that can improve the effectiveness of any unlearning method by leveraging \textit{attributes} associated with the unlearning data. These metadata have been previously ignored. 
    \item We evaluate \framework{} for two unlearning scenarios: undesired behavior (toxicity and hate speech), and hazardous scientific knowledge (biosecurity and cybersecurity). 
    %For toxicity and hate speech, we configure \framework{} to leverage demographic information (e.g., gender, ethnicity, religion, etc.). For hazardous knowledge, we configure \framework{} to leverage the scientific domain (e.g., biosecurity, cybersecurity).
    %two types of attributes: demographic information (e.g., gender, ethnicity, religion) and type of toxicity (implicit versus explicit). For hazardous knowledge, we choose the scientific domain (e.g., biosecurity, cybersecurity) to be the attribute.
    %\item 
    We empirically demonstrate that \framework\ significantly improves the performance of two recent unlearning methods on state-of-the-art LLMs (\llama\ and \zephyr). 
    %Specifically, 
    \item 
    % Compared to baselines, \framework\ reduces the percentage of toxic text generated on ToxiGen \cite{hartvigsen2022toxigen} by up to 32\%, and reduces the disclosure of hazardous biosecurity knowledge by 11.8\% and cybersecurity knowledge by 4\% on the WMDP benchmark \cite{li2024wmdp}. 
    \framework{} boosts the performance of existing unlearning techniques by up to 32\% in reducing the percentage of toxic text generated on ToxiGen \cite{hartvigsen2022toxigen}, by 11.8\%  in removing hazardous biosecurity knowledge, and by 4\% in removing hazardous cybersecurity knowledge measured on the WMDP benchmark \cite{li2024wmdp}. At the same time, \framework{} maintains general capabilities of the LLMs, measured on 10 standard academic benchmarks. 
    % \textcolor{blue}{it is not clear if this is a relative improvement, it just lists the improvement of ours, not the one of the others. Can we say XXX average reduction.? }
\end{itemize}
% However, prior work has shown limited effectiveness to \textit{unlearn} harmful behaviors and hazardous knowledge \textit{post hoc} reducing toxicity . However, these prior work mainly focus on explicit toxicity reduction, and do not measure how the toxicity reduction techniques impact model performance on benchmark tasks. 

\section{\framework\ Framework}
\label{sec:framework}

The proposed \framework{} framework is illustrated in Figure~\ref{fig:framework} and in Algorithm~\ref{alg:framework}. We focus on unlearning behaviors or bodies of knowledge (as opposed to smaller, discrete units of information) from a given LLM with parameters $\theta_{\textrm{init}}$; this is represented by a dataset $D$ consisting of examples of the undesired behavior or knowledge. We consider scenarios in which the dataset can be partitioned into subsets corresponding to different values $a_1,\dots,a_n$ of an attribute $a$ in the data which can often be identified. In the case of toxicity, for example, the attribute could be the demographic group (e.g.,~women, Muslims) targeted by the toxic text. 

Given a dataset and attribute as described above, the \framework{} framework consists of the following steps: (1) Split the dataset into subsets $D_t$ for $t = 1,\dots,n$ based on the attribute. (2) Perform unlearning separately on each subset $D_t$, all starting from the given LLM, $\theta_{\textrm{init}}$, and yielding $n$ different unlearned LLMs, $\theta^u_t$. (3) Merge the unlearned LLMs into a single final unlearned LLM, $\theta^u$. 
% \textcolor{blue}{what about adding a bit of detail about how this gets merged? It feels like we are missing technical details about this}
% \TODO{need to discuss data processing $\proc$, maybe here?} 

\begin{algorithm}[t]
   \caption{\framework $\:$ Framework} 
   \label{alg:framework}
\begin{algorithmic}
   \STATE {\bfseries Input:} Initial model parameters 
   $\theta_{\textrm{init}}$, Unlearning dataset $D$, Attribute with values $a_1, \dots, a_n$, Processing pipeline $\proc$, Unlearning method $\mathcal{U}$, Merging method $\mathcal{M}$
   \STATE {\bfseries Output:} Unlearned model $\theta^u$
   \FOR{$t=1$ {\bfseries to} $n$ }
   \STATE Select subset associated with data attribute value $a_t$ as
   \STATE $\quad D_t = \{\mathbf{x}\in D \mid \attr(\mathbf{x}) = a_t\}$
   \STATE Process subset for unlearning  
   \STATE $\quad D^u_t = \{\proc(\mathbf{x}) \mid \mathbf{x}\in D_t\}$
   \STATE Perform unlearning $\theta_t^u \leftarrow \mathcal{U}(\theta_{\textrm{init}}, D^u_t)$
   \ENDFOR
   \STATE Perform merging $\theta^u \leftarrow \mathcal{M}(\theta^u_1, \dots, \theta^u_n)$
\end{algorithmic}
\end{algorithm}

\framework{} can be instantiated with any unlearning method $\mathcal{U}(\theta_{\textrm{init}}, D^u_t)$ and merging method $\mathcal{M}(\theta^u_1,\dots,\theta^u_n)$, where the unlearning method updates model parameters from $\theta_{\textrm{init}}$ to $\theta^u_t$ using data subset $D^u_t$, and the merging method combines these independent parameters $\theta^u_1,\dots,\theta^u_n$ into one $\theta^u$. See Section~\ref{sec:instantiation} for details. 

It is frequently the case for unlearning samples to have associated attributes.  
\framework{} can be applied to a variety of attributes. 
%generality of our framework, we do not restrict Algorithm~\ref{alg:framework} to any particular attributes. 
For this reason, in Algorithm~\ref{alg:framework}, we consider a function $\attr(\cdot)$ that can output the value of a given attribute for a data sample. In practice, such a function can be implemented by using data annotations  or appropriate classifiers (e.g., a domain classifier). Similarly, we generalize any processing required by the unlearning method with function $\proc(\cdot)$. This processing function abstracts steps such as selecting representative samples with undesirable behavior or knowledge; it can also augment the unlearning set with samples of desirable behavior or knowledge to be retained.

% Section~\ref{sec:instantiation}. \TODO{maybe make it a subsection of this section}

\subsection{Instantiating \framework}
\label{sec:instantiation}
% \textbf{Instantiating} \framework:
% can be used with multiple unlearning. We have selected TA and 
We describe the specific unlearning and merging methods used in this work in the following.

\textbf{Unlearning via Task Vector Negation (\TVN)} \cite{ilharco2023editing,zhang2023composing}: This method uses the notion of \textit{task vector arithmetic} for unlearning \cite{ilharco2023editing}. Let $\theta_{\textrm{init}}\in\mathbb{R}^d$ denote the initial model weights and $\theta_{\textrm{ft}}\in\mathbb{R}^d$ the corresponding weights after fine-tuning the model on unlearning dataset $D$. The task vector used for unlearning is computed as $\tau = \theta_{\textrm{ft}} - \theta_{\textrm{init}}$. \TVN{} obtains the unlearned model as $\theta^u = \theta_{\textrm{init}} - \lambda\tau$ where $\lambda \geq 0$ is a scaling parameter. Following \citet{zhang2023composing}, we employ Parameter-Efficient Fine-Tuning (PEFT) instead of full fine-tuning and compute the task vector using Parameter Efficient Modules (PEMs). In our experiments, we use a state-of-the-art PEFT method, LoRA \cite{hu2022lora}, and perform negation using LoRA modules with $\lambda = 1$.

\textbf{Representation Misdirection Unlearning (\RMU)} \cite{li2024wmdp}: This method randomizes model activations on unlearning data while preserving model activations on data to be kept. Specifically, \RMU{} uses a two-part loss function: (1) a forget loss to bring the model activations on unlearning data close to a scaled uniform random vector, and (2) a retain loss to preserve model activations on data to be retained. Here, let $D$ denote the unlearning dataset and $D'$ denote the retain set (containing samples with desirable behavior or knowledge). Let $f_{\theta}(\cdot)$ and $f_{\theta_{\textrm{init}}}(\cdot)$ denote the hidden states of the model being unlearned and the initial model, respectively, at some layer $\ell$. Then, the forget loss is $\mathcal{L}_u = \mathbb{E}_{\mathbf{x}_u\sim D}\left[\frac{1}{|\mathbf{x}_u|}\sum_{\textrm{token }t\in\mathbf{x}_u}\left\lVert f_{\theta}(t) - c\cdot\mathbf{u}\right\rVert_2^2\right]$, where  $\mathbf{u}$ is a  random unit vector with entries sampled independently, and uniformly at random from $[0,1)$, and $c$ is a hyperparameter. The retain loss is  $\mathcal{L}_r = \mathbb{E}_{\mathbf{x}_r\sim D'}\left[\frac{1}{|\mathbf{x}_r|}\sum_{\textrm{token }t\in\mathbf{x}_r}\left\lVert f_{\theta}(t) - f_{\theta_{\textrm{init}}}(t)\right\rVert_2^2\right]$. The model parameters are updated to minimize the combined loss $\mathcal{L} = \mathcal{L}_u + \alpha\mathcal{L}_r$, where $\alpha>0$ is a hyperparameter. See Algorithm~\ref{alg:spunge-rmu-ties} in Appendix~\ref{sec:details} for additional details.
% The loss is typically computed only on layer $\ell$ and gradients are updated only on layers $\ell-2$, $\ell-1$, and $\ell$. 
% \label{sec:TA}

\textbf{\TIES-Merging} \cite{yadav2023ties}: This method allows one to merge multiple model parameters using task vector arithmetic. Given a set of model weights $\theta^u_1,\dots,\theta^u_n$ along with the initial weights $\theta_{\textrm{init}}$, \TIES-Merging computes a task vector for each model as $\tau_t = \theta^u_t - \theta_{\textrm{init}}$. Then, it operates in three steps: (i) trim each task vector by selecting the parameters with largest magnitudes,  (ii) resolve sign conflicts by creating an aggregate elected sign vector, and (iii) average only the parameters whose signs are the same as the aggregated elected sign. See Algorithm~\ref{alg:spunge-rmu-ties} in Appendix~\ref{sec:details} for additional details.

\section{Unlearning Toxicity and Hate Speech}
\label{sec:toxiciy}

\subsection{Experimental Setup}
\label{sec:setup-toxicity}
We now focus on reducing toxicity and hate speech generated by LLMs. We consider a similar experimental setup to \citet{touvron2023llama, mukherjee2023orca}. 

\textbf{Benchmarks:} To evaluate the amount of toxicity and hate speech in model generations, we use the ToxiGen benchmark \cite{hartvigsen2022toxigen}. ToxiGen is designed to measure implicit toxicity and hate speech across 13 demographic groups (e.g., African Americans, women, Mexicans, etc.). We prompt the model for completions, with toxic and benign examples from the (annotated) test subset of ToxiGen. 
Following \citet{touvron2023llama}, to measure the toxicity of the model completions, 
we use a RoBERTA model fine-tuned on ToxiGen \cite{hartvigsen2022toxigen}. We use greedy decoding and compute the percentage of completions that are deemed toxic by the classifier.

In conjuction to measuring the toxicity after unlearning, we also assess how unlearning impacts the fluency of the model. Similar to \citet{liu-etal-2021-dexperts,lu2022quark}, we measure the fluency of the outputs by computing their perplexity with an independent, larger model, \textsc{Llama2-13b}.

% (In Appendix~\ref{}, we conduct experiments with stochastic decoding.)  

To measure the general capability of the model, we consider 10 standard academic benchmarks, including all 6 benchmarks from the Open LLM Leaderboard \cite{open-llm-leaderboard}. See Appendix~\ref{sec:benchmarks} for the list of benchmarks. We perform evaluations using the Language Model Evaluation Harness framework~\cite{gao2023lm-eval}.

\textbf{Unlearning Dataset:} We used the annotated training subset of ToxiGen consisting of of 8,960 samples of both benign and toxic examples across 13 demographic groups. 
% The annotations contain among others the target demographic group for the prompt, toxicity levels (between 1 to 4) from human annotators.  
% \textcolor{blue}{let's add a reference that uses this way of measuring toxicity.}

\subsection{\framework\ Leveraging Demographic Information}
\label{sec:spunge-demographics}
We instantiate \framework\ using the demographic information in the unlearning set as attributes. We use the ToxiGen subset for unlearning which contains, for each prompt, the target demographic group and the toxicity level evaluated by human annotators. While ToxiGen encompasses 13 demographic groups, for our experiments, we choose the following 5 representative demographic groups: Nationality (Mexican), Gender and Sex (Women), Religion (Muslim), Sexual Orientation (LGBTQ), and Health Condition (Physical Disability). 
 
\framework\ first splits the unlearning set into 5 subsets -- $D_1, \dots, D_5$ -- each associated with one of the above 5 demographic groups. Next, from each set $D_t$, we select a subset of 
% data with high toxicity. For this, we use the toxicity scores annotated by humans, and select 
samples for which the toxicity level $\geq$ 3.
% \footnote{The range of toxicity annotation scores is from 1 to 5.} 
This yields five unlearning subsets $D_1^u, \dots, D_5^u$. \framework\ then performs unlearning on the base model $\theta_{\textrm{init}}$ with each $D_t^u$ to obtain $\theta_1^u, \dots, \theta_5^u$. Finally, we use \TIES-merging (Section~\ref{sec:instantiation}) to merge the five unlearned models.

\subsection{Experimental Results}
\label{sec:results-toxicity}
We perform unlearning on two state-of-the-art models, \zephyr{} \cite{tunstall2023zephyr} and \llama{} \cite{touvron2023llama}. We consider \RMU{} and \TVN{} (Section~\ref{sec:instantiation}) as the unlearning methods and instantiate \framework{} with each. 
Table~\ref{tab:toxicity} shows results for the two LLMs and two unlearning methods. 
The most relevant comparisons are between an unlearning method (\RMU{} or \TVN{}) and its \framework-enhanced version.
% \RMU{} reduces the percentage of toxicity in the model generations on ToxiGen from 20.48 to 14.61 with slight increase in the perplexity.
With \zephyr, \framework\ boosts the performance of both \RMU{} and \TVN. Specifically,  
\framework{} reduces the toxicity percentage of \RMU{} by 32\% (from 14.61 to 9.89) and of \TVN{} by 31\% (from 5.65 to 3.88), while maintaining the fluency of generations as measured by the perplexity computed with \textsc{Llama-13b}. Notably, \framework{} maintains general capabilities of the model as measured by the average accuracy on the benchmarks. Similarly, for \llama, \framework\ reduces the toxicity percentage of \TVN{} by 30\% (from 4.26 to 2.96) while maintaining the average accuracy on benchmarks within 1\% of the base model\footnote{We have so far been unable to obtain satisfactory results with \RMU{} for \llama, since we found it tricky to tune RMU's hyperparameters for \llama\ and \citet{li2024wmdp} did not provide guidance on this. For \RMU{} with \zephyr, we use the hyperparameters from \citet{li2024wmdp}.}.
In Appendix~\ref{sec:toxicity-per-demographic}, we compare the toxicity percentage for each demographic and show that \framework{} strengthens the baseline methods. In Appendix~\ref{sec:toxicity-type}, we instantiate \framework{} to leverage the attribute of type of toxicity.

\begin{table}[t]
\caption{\textbf{Evaluation of toxicity unlearning on ToxiGen.} Toxicity is the percentage of toxic generations, PPL is the perplexity of generations measured with \textsc{Llama2-13b}, and Average Acc. is the average performance on 10 benchmarks (Appendices~\ref{sec:benchmarks} and~\ref{sec:detailed-results}). \framework{} is configured to leverage demographic information.}
\label{tab:toxicity}
\vskip 0.1in
\centering
\begin{sc}
\resizebox{\columnwidth}{!}{%
\begin{tabular}{lccc}
\toprule
{Model} & \multicolumn{2}{c}{ToxiGen} & Average \\
{+ Method} & Toxicity ($\downarrow$) & PPL ($\downarrow$) & Acc. ($\uparrow$)\\
\midrule
\zephyr & 20.48 & 7.62 & 65.72 \\
% \midrule
+ \RMU & 14.61  & 8.05  & 65.92\\
+ \framework-\RMU   &  \textbf{9.89} & \textbf{8.03} &  \textbf{65.97}\\
% \midrule
+ \TVN & 5.65  & \textbf{8.36}  & \textbf{65.67}\\
+ \framework-\TVN   &  \textbf{3.88} & {8.66} &  {65.53}\\
\midrule
% \midrule
\llama & 15.95 & 5.97 & 56.29\\
% \midrule
+ \TVN    & 4.26 & 8.42 & \textbf{56.35}\\
+ \framework-\TVN & \textbf{2.96}  & \textbf{7.88} &   55.72\\ 
\bottomrule
\end{tabular}
}
\end{sc}
\vskip -0.1in
\end{table}

\begin{table}[t]
\caption{\textbf{Evaluation of hazardous knowledge unlearning on WMDP.} \framework{} strengthens the performance of \RMU, while preserving general capabilities of the model.}
\label{tab:wmdp}
\vskip 0.1in
\centering
\begin{sc}
\resizebox{\columnwidth}{!}{%
\begin{tabular}{lccc}
\toprule
Model & WMDP-Bio  & WMDP-Cyber & MMLU \\
{+ Method} & ($\downarrow$) & ($\downarrow$) & ($\uparrow$)\\
\midrule
\zephyr & 63.55 & 43.63 & 58.15 \\
+ \RMU & 31.26  & 27.62  & \textbf{56.48}\\
+ \framework-\RMU   &  \textbf{27.57} & \textbf{26.47} &  {55.83}\\
\bottomrule
\end{tabular}
}
\end{sc}
% \vskip -0.1in
\end{table}

\section{Unlearning Hazardous Knowledge}
\label{sec:knowledge}

\subsection{Experimental Setup} 
\label{sec:knowledge-setup}
We now focus on reducing the model's ability to answer questions about hazardous knowledge (e.g., cultivating virus) while maintaining the ability to answer questions about non-hazardous knowledge (e.g., properties of fungi).  We follow the experimental setup of \citet{li2024wmdp}. 

\textbf{Benchmarks:} To evaluate hazardous knowledge removal, we use the Weapons of Mass Destruction Proxy (WMDP) benchmark \cite{li2024wmdp} which consists of 3,668 multiple-choice questions on biosecurity (WMDP-Bio), cybersecurity (WMDP-Cyber), and chemistry (WMDP-Chem). To evaluate general-knowledge question answering, we use the Massive Multitask Language Understanding (MMLU) benchmark \cite{hendrycks2021measuring}. Similar to \citet{li2024wmdp}, we conduct unlearning experiments only on the challenging subsets WMDP-Bio and WMDP-Cyber. We again evaluate using %perform evaluations using 
the Language Model Evaluation Harness framework~\cite{gao2023lm-eval}.

\textbf{Unlearning Dataset:} For unlearning, we use the \textit{bio corpora} and \textit{cyber corpora} collected by \citet{li2024wmdp} and released publicly.
% ~\footnote{https://github.com/centerforaisafety/wmdp}. 
The bio corpora consist of a selected subset of PubMed papers that are related to the topics appearing in WMDP-Bio questions. The cyber corpora consist of passages scraped from GitHub via keyword search on topics related to WMDP-Cyber questions.  

\textbf{Baseline:} We consider \RMU{} (Section~\ref{sec:instantiation}) as the baseline unlearning method. \RMU{} has been shown to be superior to several unlearning methods for hazardous knowledge unlearning \cite{li2024wmdp}. In our preliminary experiments, \TVN{} (Section~\ref{sec:instantiation}) was not able to successfully unlearn hazardous knowledge while retaining general performance.

% \textbf{Models:} We perform unlearning from two state-of-the-art models \textsc{Zephyr-7b-beta} \cite{tunstall2023zephyr} and {\textcolor{purple}{\textsc{Llama2-7b}}}.

\subsection{\framework\ Leveraging Scientific Domains}
\label{sec:spunge-knowledge}
We instantiate \framework{} to leverage the scientific domain attribute in the unlearning set. 
As mentioned in the previous section, the unlearning dataset is a combination of bio and cyber corpora. We split the data by domain to separate bio corpora ($D_1$) and cyber corpora ($D_2$). \framework{} performs unlearning separately on each of them to obtain two unlearned LLMs: one with biosecurity hazardous knowledge removed $\theta^u_1$ and the other with cybersecurity hazardous knowledge removed $\theta^u_2$. \framework{} then merges $\theta^u_1$ and $\theta^u_2$ using \TIES-merging (Section~\ref{sec:instantiation}).
Note that, in contrast to \framework{}, \RMU{} (and other baselines) in \citet{li2024wmdp} use the bio and cyber corpora together during unlearning -- in particular, \RMU{} alternates between one batch from the bio corpora and one from the cyber corpora during unlearning.

\subsection{Experimental Results}
Table~\ref{tab:wmdp} shows that \framework\ fortifies the performance of \RMU{} in removing hazardous knowledge while maintaining general-knowledge capabilities. In particular, \framework{} reduces WMDP-Bio accuracy by 11.8\% (from 31.26 to 27.57) and WMDP-Cyber accuracy by 4\% (from 27.62 to 26.47), while maintaining MMLU accuracy within 1\% of \RMU{}.

% \section{Discussion}
% \label{sec:discussion}

\section{Conclusion}
\label{sec:conclusion}
We presented \framework{}, a novel unlearning framework that takes advantage of attributes associated with the data to be unlearned. \framework{} can be instantiated with any unlearning method to boost its performance. \framework{} leverages attributes using a \textit{split-unlearn-then-merge} approach. We considered two unlearning scenarios: unlearning undesirable behavior (i.e., toxicity) and hazardous knowledge (i.e., biosecurity and cybersecurity). Through our experiments, we demonstrated that \framework{} significantly improves the effectiveness of two state-of-the-art unlearning methods. Interesting future works would explore using \framework{} for data unlearning (e.g., copyrighted or licensed data).

\bibliography{example_paper}
\bibliographystyle{icml2024}

%%%%%%%%%%%%%%%%%%%%%%%%%%%%%%%%%%%%%%%%%%%%%%%%%%%%%%%%%%%%%%%%%%%%%%%%%%%%%%%
%%%%%%%%%%%%%%%%%%%%%%%%%%%%%%%%%%%%%%%%%%%%%%%%%%%%%%%%%%%%%%%%%%%%%%%%%%%%%%%
% APPENDIX
%%%%%%%%%%%%%%%%%%%%%%%%%%%%%%%%%%%%%%%%%%%%%%%%%%%%%%%%%%%%%%%%%%%%%%%%%%%%%%%
%%%%%%%%%%%%%%%%%%%%%%%%%%%%%%%%%%%%%%%%%%%%%%%%%%%%%%%%%%%%%%%%%%%%%%%%%%%%%%%
\newpage
\appendix
% \onecolumn
\section{Related Work}
\label{sec:related-work}

\textbf{Machine Unlearning:} The notion of machine unlearning was first introduced by \citet{cao2015towards} motivated by the \textit{right-to-be-forgotten} and focused on removing specific training samples. Since then, there have been a number of works that have focused on removing specific training data samples via unlearning~\cite{bourtoule2021machine,graves2020amnesiac,izzo2021approximate,ginart2019making,golatkar2020eternal,golatkar2020forgetting,thudi2021unrolling}. 
% and surveys \cite{nguyen2022survey,xu2023machine}. 

Unlearning for LLMs has started to gain recent attention resulting in works in data unlearning \cite{jang2023knowledge,wang2023kga,kassem2023preserving,maini2024tofu,zhang2024negative}, concept unlearning \cite{eldan2023whos}, behavior unlearning \cite{lu2022quark,yao2024large,liu2024safer}, knowledge unlearning \cite{li2024wmdp}. Recent surveys have shown additional methods where unlearning has been applied \cite{nguyen2022survey,xu2023machine,liu2024rethinking}.
% \TODO{in progress, mostly needs to be moved to appendix for space} Almost all the prior work on unlearning in LLMs has focused on developing efficient unlearning methods, assuming that the data used for unlearning is  given \cite{nguyen2022survey,xu2023survey}.
Prior works have mainly focused on designing unlearning methods, evaluation metrics, and benchmarks. However, they do not take into account attributes of data used for unlearning. Our proposed \framework{} leverages data attributes to fortify the performance of any unlearning method.

\textbf{Toxicity Reduction in LLMs:} Early works in reducing toxicity in language models \cite{krause2021gedi,liu-etal-2021-dexperts,dathathri2020Plug} have focused on small to moderated sized models and restrict to explicit toxicity. Detoxification techniques primarily employ controlled text generation methods, which incurs heavy inference overhead and it is difficult to measure model performance on benchmark tasks. Machine unlearning provides an alternative for mitigating toxicity in LLMs \cite{ilharco2023editing,zhang2023composing,lu2022quark}.

\section{Benchmarks Used for Evaluation}
\label{sec:benchmarks}
We use the following 10 benchmarks, including all 6 from Open LLM Benchmark \cite{open-llm-leaderboard}. For the benchmarks included in Open LLM Benchmark, we select the same number of shots as prescribed. For the remaining 4 benchmarks, we perform 5-shot prompting.
\begin{itemize}
    \item AI2 Reasoning Challenge (ARC-Challenge and ARC-Easy) \cite{clark2018think} (25-shot)
    \item HellaSwag \cite{zellers2019hellaswag} (10-shot)
    \item MMLU \cite{hendrycks2021measuring} (5-shot)
    \item TruthfulQA \cite{lin2022truthfulqa} (0-shot)
    \item Winogrande \cite{sakaguchi2021winogrande} (5-shot)
    \item GSM8K \cite{cobbe2021training} (5-shot)
    \item MathQA \cite{amini2019mathqa} (5-shot)
    \item PIQA \cite{bisk2019piqa} (5-shot)
    \item PubMedQA \cite{jin2019pubmedqa} (5-shot)
\end{itemize}

\section{Experimental Results: Details and Additional Results}
\label{sec:detailed-results}

\subsection{Experiment Details}
\label{sec:details}

\textbf{\framework{} with \RMU{} and \TIES:} Algorithm~\ref{alg:spunge-rmu-ties} presents the instantiation of \framework{} with \RMU{} and \TIES.

\textbf{\RMU{} with \zephyr:} We use the hyperparameters from \cite{li2024wmdp}. In particular, we use $c = 6.5$ and $\alpha = 1200$. We use a learning rate of $5 \times 10^{-5}$ and a batch size of 150 with the Adam optimizer. We select layer 7 to perform the unlearning loss and layers 5, 6, and 7 to update gradients. When performing separate unlearning with \framework{}, the unlearning subsets are substantially smaller. Thus, we perform training for 2 epochs with early stopping if the cosine similarity between the activations of the unlearned model and the initial model drops below 0.5.

\textbf{\TVN{} with \zephyr:} We set the LoRA rank to 16, $\alpha$ associated with LoRA to 16, LoRA dropout to 0.01, and target modules as the default modules in HuggingFace PEFT library. We use a learning rate of $2 \times 10^{-5}$ with the Adam optimizer and cosine learning rate schedule to train for 1 epoch. When performing separate unlearning with \framework{}, the unlearning subsets are substantially smaller. Thus, we perform training with a learning rate of $1 \times 10^{-4}$ for 1 epoch.

\textbf{\TVN{} with \llama:} We set the LoRA rank to 64, $\alpha$ associated with LoRA to 64, LoRA dropout to 0.01, and target modules as \texttt{key}, \texttt{value}, \texttt{query}, \texttt{up}, \texttt{down}, and \texttt{gate} projections. We set learning rate to be $1 \times 10^{-4}$ with the Adam optimizer and cosine learning rate scheduling.

\textbf{Performance on Academic Benchmarks:} We present the performance on 10 academic benchmarks (Appendix~\ref{sec:benchmarks}) in Tables~\ref{tab:zephyr-rmu}, \ref{tab:zehyr-tvn}, and \ref{tab:llama-tvn}.

\subsection{Toxicity per Demographic Group}
\label{sec:toxicity-per-demographic}
\begin{figure}[tbp]
% \vskip 0.2in
\begin{center}
\centerline{\includegraphics[width=.85\columnwidth]{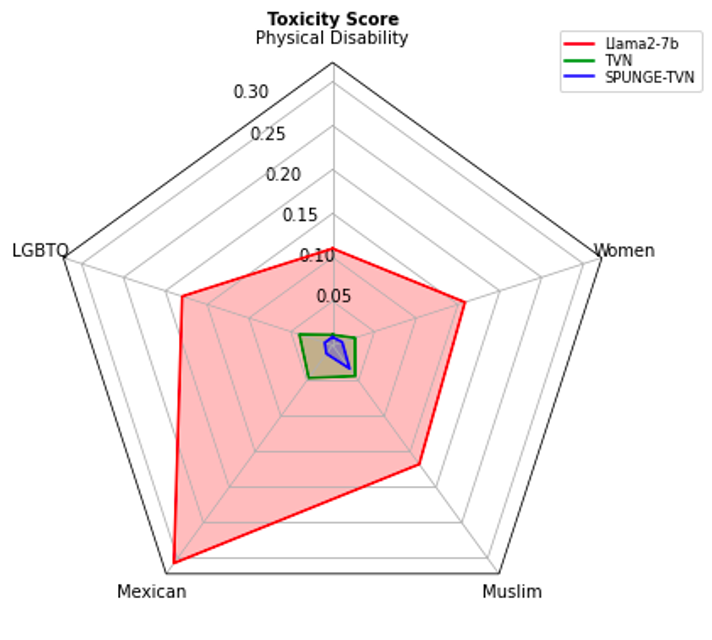}}
\caption{Toxicity scores per demographic group on ToxiGen test set for the \llama\ base model, after unlearning with \TVN, and after unlearning with \framework\ used with \TVN.}
\label{fig:toxicity-demographic-llama-tvn}
\end{center}
\vskip -0.2in
\end{figure}

\begin{figure}[tbp]
% \vskip 0.2in
\begin{center}
\centerline{\includegraphics[width=.85\columnwidth]{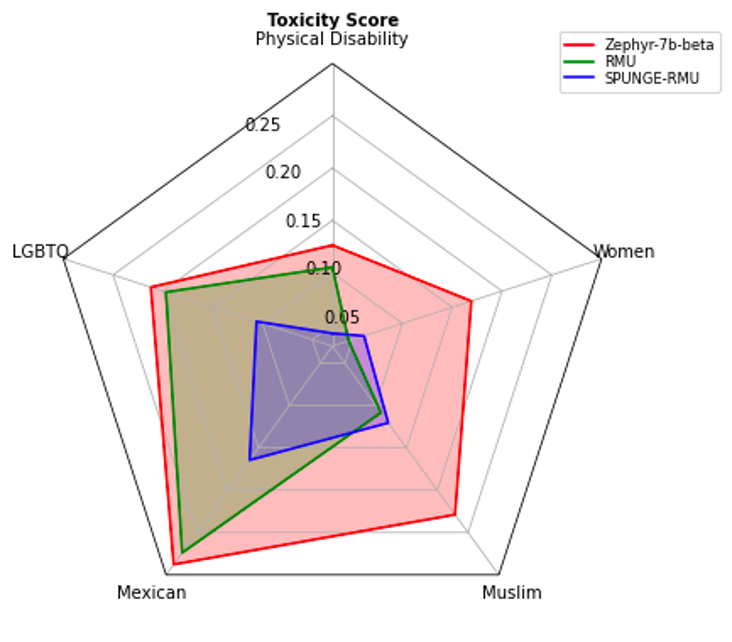}}
\caption{Toxicity scores per demographic group on ToxiGen test set for the \zephyr\ base model, after unlearning with \RMU, and after unlearning with \framework\ used with \RMU.}
\label{fig:toxicity-demographic-zehpyr-rmu}
\end{center}
\vskip -0.2in
\end{figure}

\begin{figure}[tbp]
% \vskip 0.2in
\begin{center}
\centerline{\includegraphics[width=.85\columnwidth]{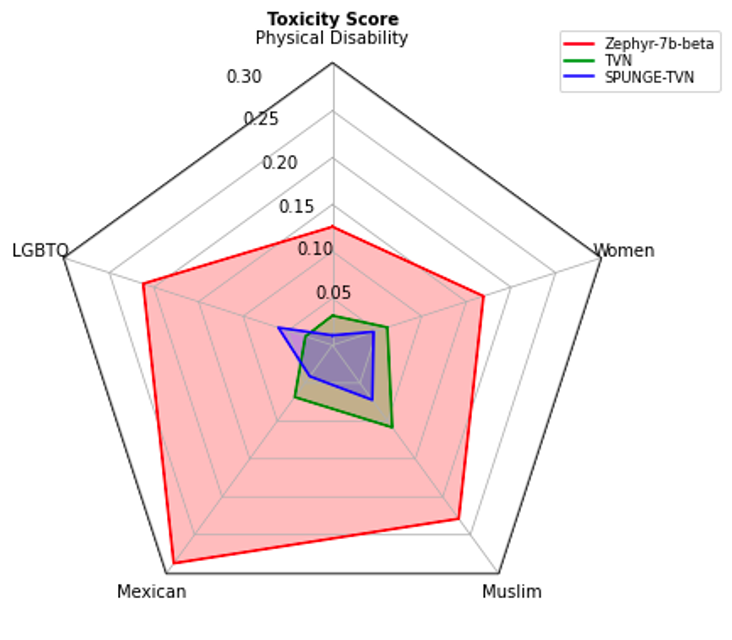}}
\caption{Toxicity scores per demographic group on ToxiGen test set for the \zephyr\ base model, after unlearning with \TVN, and after unlearning with \framework\ used with \TVN.}
\label{fig:toxicity-demographic-zehyr-tvn}
\end{center}
\vskip -0.2in
\end{figure}

We analyze the percentage of toxic generations for each demographic group. We focus on the same 5 demographic groups used during unlearning: Nationality (Mexican), Gender and Sex (Women), Religion (Muslim), Sexual Orientation (LGBTQ), and Health Condition (Physical Disability). In Figures~\ref{fig:toxicity-demographic-llama-tvn}, \ref{fig:toxicity-demographic-zehpyr-rmu}, and \ref{fig:toxicity-demographic-zehyr-tvn}, we present radar plots for toxicity percentage per demographic group. The plots show results for the base model, a baseline unlearning, and \framework{} used with the baseline. \framework{} reduces the toxicity for every demographic group for \llama\ (Figure~\ref{fig:toxicity-demographic-llama-tvn}) whereas for \zephyr, \framework{} cuts down toxicity percentage for most demographic groups (Figures~\ref{fig:toxicity-demographic-zehpyr-rmu} and \ref{fig:toxicity-demographic-zehyr-tvn}).

\begin{table}[tbp]
\caption{Accuracy of the benchmarks for the \zephyr\ model and the models after performing unlearning on ToxiGen.}
\label{tab:zephyr-rmu}
\vskip 0.1in
\centering
\begin{sc}
\resizebox{\columnwidth}{!}{%
\begin{tabular}{lccc}
\toprule
{Benchmark} & \zephyr & \RMU & \framework\\ 
\midrule
Arc-C ($\uparrow$) & 63.90 & 63.31 & 63.65 \\ 
Arc-E ($\uparrow$) & 84.89 & 84.93 & 84.89\\ 
HellaSwag ($\uparrow$) & 84.21 & 84.16 & 84.14 \\ 
MMLU ($\uparrow$) & 59.75 & 59.82 & 59.73 \\ 
Winogrande ($\uparrow$) & 77.42 & 78.13 & 77.82\\ 
GSM8K ($\uparrow$) & 34.42 & 34.64 & 34.87\\ 
MathQA ($\uparrow$) & 38.05 & 37.82 & 38.35\\ 
PIQA ($\uparrow$) & 82.69 & 82.91 & 82.75\\ 
PubmedQA ($\uparrow$) & 76.80 & 77.00 & 76.60\\ 
TruthfulQA ($\uparrow$) & 55.12 & 56.52 & 56.92\\
\midrule
Average ($\uparrow$) & 65.72 & 65.92 & 65.97\\
\bottomrule
\end{tabular}
}
\end{sc}
\vskip -0.1in
\end{table}

\begin{table}[tbp]
\caption{Accuracy on the benchmarks for the \zephyr\ model and the models after performing unlearning on ToxiGen.}
\label{tab:zehyr-tvn}
\vskip 0.1in
\centering
\begin{sc}
\resizebox{\columnwidth}{!}{%
\begin{tabular}{lccc}
\toprule
{Benchmark} & \zephyr & \TVN & \framework\\ 
\midrule
Arc-C ($\uparrow$) & 63.90 & 64.50 & 63.73 \\ 
Arc-E ($\uparrow$) & 84.89 & 83.96 & 83.37\\ 
HellaSwag ($\uparrow$) & 84.21 & 84.41 & 84.28 \\ 
MMLU ($\uparrow$) & 59.75 & 58.14 & 58.52\\ 
Winogrande ($\uparrow$) & 77.42 & 78.05 & 77.82\\ 
GSM8K ($\uparrow$) & 34.42 & 34.79 & 33.43\\ 
MathQA ($\uparrow$) & 38.05 & 36.88 & 36.71\\ 
PIQA ($\uparrow$) & 82.69 & 8226 & 82.42\\ 
PubmedQA ($\uparrow$) & 76.80 & 76.60 & 77.00\\ 
TruthfulQA ($\uparrow$) & 55.12 & 57.20 & 58.01\\
\midrule
Average ($\uparrow$) & 65.72 & 65.67 & 65.52\\
\bottomrule
\end{tabular}
}
\end{sc}
\vskip -0.1in
\end{table}

\begin{table}[!tbp]
\caption{Accuracy on the benchmarks for the \llama\ model and the models after performing unlearning on ToxiGen.}
\label{tab:llama-tvn}
\vskip 0.1in
\centering
\begin{sc}
\resizebox{.95\columnwidth}{!}{%
\begin{tabular}{lcccccccccc}
\toprule
{Benchmark} & \llama & \TVN & \framework\\ 
\midrule
Arc-C ($\uparrow$) & 53.32 & 53.32 & 52.04\\ 
Arc-E ($\uparrow$) & 81.48 & 81. 64 & 81.69\\ 
HellaSwag ($\uparrow$) & 78.57 & 77.44 & 74.39\\ 
MMLU ($\uparrow$) & 45.99 & 44.74 & 44.22\\ 
Winogrande ($\uparrow$) & 72.45 & 73.71 & 74.11\\ 
GSM8K ($\uparrow$) & 15.01 & 8.11 & 9.47\\ 
MathQA ($\uparrow$) & 29.41 & 29.31 & 29.14\\ 
PIQA ($\uparrow$) & 79.37 & 79.97 & 79.65\\ 
PubmedQA ($\uparrow$) & 68.40 & 71.00 & 69.80\\ 
TruthfulQA ($\uparrow$) & 38.97 & 44.34 & 42.72\\
\midrule
Average ($\uparrow$) & 56.29 & 56.35 & 55.72\\
\bottomrule
\end{tabular}
}
\end{sc}
\vskip -0.1in
\end{table}

\begin{algorithm*}[tb]
   \caption{\framework $\:$ Framework Instantiated with \RMU{} \cite{li2024wmdp} and \TIES-Merging \cite{yadav2023ties}}
   \label{alg:spunge-rmu-ties}
\begin{algorithmic}
   \STATE {\bfseries Input:} Initial model parameters $\theta_{\textrm{init}}$, Dataset $D$ for unlearning, Retain dataset $D^r$ (as needed by \RMU), Data attributes $a_1, \dots, a_n$, Parameters for \RMU\ $c$, $\alpha$, Parameters for \TIES-merging $\lambda,k$
   \STATE {\bfseries Output:} Unlearned model $\theta_u$
   \FOR{$t=1$ {\bfseries to} $n$ }
   \STATE Select subset associated with data attribute $a_t$ as
   $D_t = \{\mathbf{x}\in D \mid \attr(\mathbf{x}) = a_t\}$
   \STATE Process subset for unlearning  
    $D^u_t = \{\proc(\mathbf{x}) \mid \mathbf{x}\in D_t\}$
   \STATE Perform unlearning $\theta_i^u \leftarrow \RMU(\theta_{\textrm{init}}, D^u_t, D^r,c,\alpha)$
   \ENDFOR
   \STATE Perform merging $\theta^u \leftarrow \TIES(\theta^u_1, \dots, \theta^u_n,\theta_{\textrm{init}},\lambda)$
    \STATE
   \STATE {\bfseries Function} $\RMU(\theta,D^u,D^r,c,\alpha)$
   \STATE Sample unit vector $\mathbf{u}$  with entries drawn independently, and uniformly at random from $[0, 1)$ 
   \FOR{data points $\mathbf{x}_u \sim D^u$, $\mathbf{x}_r \sim D^r$}
   \STATE Set $\mathcal{L}_u = \frac{1}{L}\sum_{t\in\mathbf{x}_u}\left\lVert f_{\theta}(t) - c\cdot\mathbf{u}\right\rVert_2^2$, where $\mathbf{x}_u$ contains $L$ tokens
   \STATE Set $\mathcal{L}_r = \frac{1}{L}\sum_{t\in\mathbf{x}_r}\left\lVert f_{\theta}(t) - f_{\theta_{\textrm{init}}}(t)\right\rVert_2^2$, where $\mathbf{x}_r$ contains $L$ tokens
   \STATE Update parameters $\theta$ using $\mathcal{L} = \mathcal{L}_u + \alpha \cdot \mathcal{L}_r$
   \ENDFOR
   \STATE {\bfseries return} $\theta$

   \STATE
   \STATE {\bfseries Function} $\TIES(\theta_1,\dots,\theta_n,\theta_{\textrm{init}},\lambda,k)$
   \FOR{$t=1$ {\bfseries to} $n$}
   \STATE Create task vector $\tau_t = \theta^u_t - \theta_{\textrm{init}}$
   \STATE Sparsify the task vector to keep only largest $k$ elements to obtain $\hat{\tau}_t$
   \STATE Collect signs for components $\hat{\gamma}_t \leftarrow \textrm{sign}(\hat{\tau}_t)$ 
   \STATE Collect magnitudes for components $\hat{\mu} \leftarrow |\hat{\tau}_t|$ 
   \ENDFOR
   \STATE Elect final signs as $\gamma_u \leftarrow \textrm{sign}\left(\sum_{t=1}^n\hat{\tau}_t\right)$
   \FOR{$p=1$ {\bfseries to} $d$}
   \STATE $\mathcal{A}^p = \{t \in [n] \mid \hat{\gamma}_t^p = \gamma^p\}$ 
   \STATE $\tau^p_u = \frac{1}{|\mathcal{A}^p|}\sum_{t\in\mathcal{A}^p}\hat{\tau}^p_t$ 
   \ENDFOR
    \STATE $\theta_u \leftarrow \theta_{\textrm{init}} + \lambda\tau_u$
   \STATE  {\bfseries return} $\theta_u$
\end{algorithmic}
\end{algorithm*}

\subsection{\framework{} Leveraging Type of Toxicity}
\label{sec:toxicity-type}
We consider the goal of unlearning implicit as well as explicit toxicity from LLMs. Explicit toxicity is a conventional form of toxicity containing profanity, slurs, swearwords, and offensive language. On the other hand, implicit toxicity does not such terms in contrast to explicit toxicity and can even be positive in sentiment \cite{hartvigsen2022toxigen}. Examples of implicit toxicity include stereotyping and microaggressions. The ToxiGen dataset \cite{hartvigsen2022toxigen} is focused on implicit and subtly toxic samples. There are datasets that contains samples with explicit toxicity such as Civil Comments \cite{borkan2019civilcomments}.

As a baseline we perform unlearning on \llama{} with \TVN{} using a dataset consisting of samples with implicit as well as explicit toxicity. To represent implicit toxicity, we take samples from the (annotated) train set of ToxiGen with human toxicity level of 5 (highest level). To represent explicit toxicity, we take samples from Civil Comments with severe toxicity score greater than 0.35. We use the same hyperparameters from Section~\ref{sec:details}.

For comparison, we instantiate \framework{} to leverage type of toxicity. Specifically, we separate the unlearning set into two subsets: examples with implicit toxicity ($D_1$) and examples with explicit toxicity ($D_2$). We separately unlearn the two subsets, and them merge the unlearning models with \TIES-merging. 

Table~\ref{tab:toxicity-txgen-rtp} compares \TVN{} and its \framework-enhanced version. In addition to computing toxicity on the ToxiGen test set (which contains implicitly toxic and benign samples), we also compute toxicity on Real Toxicity Prompts (RTP) \cite{gehman2020realtoxicityprompts} (which contains explicitly toxic and benign samples). We see that \framework{} amplifies the performance of \TVN{} on both ToxiGen and RTP, while maintaining the performance on benchmark tasks. We present the accuracy results on benchmark tasks in Table~\ref{tab:llama-tvn-cc-txgen}.

\begin{table}[tbp]
\caption{Evaluation of toxicity unlearning on ToxiGen and RealToxicityPrompts (RTP). We consider \llama\ with \TVN. Toxicity is the percentage of toxic generations and Average Acc. is the average performance on the 10 benchmarks. \framework{} is configured to leverage type of toxicity: implicit versus explicit toxicity.}
\label{tab:toxicity-txgen-rtp}
\vskip 0.1in
\centering
\begin{sc}
\resizebox{\columnwidth}{!}{%
\begin{tabular}{lccc}
\toprule
{Model} & \multicolumn{2}{c}{Toxicity} & Average \\
{+ Method} & ToxiGen ($\downarrow$) & RTP ($\downarrow$) & Acc. ($\uparrow$)\\
\midrule
\llama & 15.95 & 6.40 & 56.29\\
+ \TVN    & 8.42 & 3.17 & \textbf{56.14}\\
+ \framework-\TVN & \textbf{4.81}  & \textbf{1.97} &   55.23\\ 
\bottomrule
\end{tabular}
}
\end{sc}
\vskip -0.1in
\end{table}

\begin{table}[tbp]
\caption{Accuracy on the benchmarks for the \llama\ model and the models after performing unlearning on Civil Comments and ToxiGen.}
\label{tab:llama-tvn-cc-txgen}
\vskip 0.1in
\centering
\begin{sc}
\resizebox{.95\columnwidth}{!}{%
\begin{tabular}{lcccccccccc}
\toprule
{Benchmark} & \llama & \RMU & \framework\\ 
\midrule
Arc-C ($\uparrow$) & 53.32	& 53.75 & 53.24\\ 
Arc-E ($\uparrow$) & 81.48	& 81.35 & 79.33\\ 
HellaSwag ($\uparrow$) & 78.57 & 78.41 & 77.82\\ 
MMLU ($\uparrow$) & 45.99 & 44.32 & 44.16\\ 
Winogrande ($\uparrow$) & 72.45 & 73.16 & 73.16\\ 
GSM8K ($\uparrow$) & 15.01 & 11.44 & 4.16\\ 
MathQA ($\uparrow$) & 29.41 & 29.34 & 29.41\\ 
PIQA ($\uparrow$) & 79.37 & 79.05 & 79.65\\ 
PubmedQA ($\uparrow$) & 68.40 & 70.20 & 70.20\\ 
TruthfulQA ($\uparrow$) & 38.97 & 40.40 & 41.23\\
\midrule
Average ($\uparrow$) & 56.29 & 56.14 & 55.23\\
\bottomrule
\end{tabular}
}
\end{sc}
\vskip -0.1in
\end{table}

%%%%%%%%%%%%%%%%%%%%%%%%%%%%%%%%%%%%%%%%%%%%%%%%%%%%%%%%%%%%%%%%%%%%%%%%%%%%%%%
%%%%%%%%%%%%%%%%%%%%%%%%%%%%%%%%%%%%%%%%%%%%%%%%%%%%%%%%%%%%%%%%%%%%%%%%%%%%%%%

\end{document}